\documentclass[letterpaper,twoside,twocolumn,british,english,letterpaper, 10 pt, conference]{ieeeconf}
\usepackage[T1]{fontenc}
\usepackage[latin9]{inputenc}
\usepackage{xcolor}
\usepackage{babel}
\usepackage{units}
\usepackage{mathtools}
\usepackage{amsmath}
\usepackage{amssymb}
\usepackage{graphicx}
\usepackage[pdfusetitle,
 bookmarks=true,bookmarksnumbered=false,bookmarksopen=false,
 breaklinks=false,pdfborder={0 0 0},pdfborderstyle={},backref=false,colorlinks=false]
 {hyperref}

\makeatletter

\pdfpageheight\paperheight
\pdfpagewidth\paperwidth

\providecommand{\tabularnewline}{\\}

\usepackage{graphicx}
\usepackage{import}
\usepackage{makecell}
\graphicspath{{images/}}

\usepackage{anyfontsize} 
\usepackage[font=footnotesize]{caption} 
\usepackage{url}
\usepackage{cite}

\IEEEoverridecommandlockouts                              

\ifdefined\showcaptionsetup
 \PassOptionsToPackage{caption=false}{subfig}
\fi
\usepackage{subfig}
\makeatother

\begin{document}
\global\long\def\dq#1{\underline{\boldsymbol{#1}}}%

\global\long\def\quat#1{\boldsymbol{#1}}%

\global\long\def\mymatrix#1{\boldsymbol{#1}}%

\global\long\def\myvec#1{\boldsymbol{#1}}%

\global\long\def\mapvec#1{\boldsymbol{#1}}%

\global\long\def\dualvector#1{\underline{\boldsymbol{#1}}}%

\global\long\def\dual{\varepsilon}%

\global\long\def\dotproduct#1{\langle#1\rangle}%

\global\long\def\norm#1{\left\Vert #1\right\Vert }%

\global\long\def\mydual#1{\underline{#1}}%

\global\long\def\hamilton#1#2{\overset{#1}{\operatorname{\mymatrix H}}\left(#2\right)}%

\global\long\def\hamiquat#1#2{\overset{#1}{\operatorname{\mymatrix H}}_{4}\left(#2\right)}%

\global\long\def\hami#1{\overset{#1}{\operatorname{\mymatrix H}}}%

\global\long\def\tplus{\dq{{\cal T}}}%

\global\long\def\getp#1{\operatorname{\mathcal{P}}\left(#1\right)}%

\global\long\def\getd#1{\operatorname{\mathcal{D}}\left(#1\right)}%

\global\long\def\swap#1{\text{swap}\left(#1\right)}%

\global\long\def\imi{\hat{\imath}}%

\global\long\def\imj{\hat{\jmath}}%

\global\long\def\imk{\hat{k}}%

\global\long\def\real#1{\operatorname{\mathrm{Re}}\left(#1\right)}%

\global\long\def\imag#1{\operatorname{\mathrm{Im}}\left(#1\right)}%

\global\long\def\imvec{\boldsymbol{\imath}}%

\global\long\def\vector{\operatorname{vec}}%

\global\long\def\mathpzc#1{\fontmathpzc{#1}}%

\global\long\def\cost#1#2{\underset{\text{#2}}{\operatorname{\text{C}}}\left(\ensuremath{#1}\right)}%

\global\long\def\diag#1{\operatorname{diag}\left(#1\right)}%

\global\long\def\frame#1{\mathcal{F}_{#1}}%

\global\long\def\ad#1#2{\text{Ad}\left(#1\right)#2}%

\global\long\def\spin{\text{Spin}(3)}%

\global\long\def\spinr{\text{Spin}(3){\ltimes}\mathbb{R}^{3}}%

\global\long\def\argmintwo#1#2#3#4{ \begin{aligned}\myvec u_{q}  &  \in\underset{\dot{\myvec{#1}}}{\text{argmin}}  &   &  #2 \\
  &  \text{subject to}  &   &  #3 \\
  &   &   &  #4 
\end{aligned}
 }%

\global\long\def\argminfour#1#2#3#4#5#6{ \begin{aligned}\myvec u_{\hat{a}}  &  \in\underset{\dot{#1}}{\text{argmin}}  &   &  #2 \\
  &   &   &  #3 \\
  &  \text{subject to}  &   &  #4 \\
  &   &   &  #5 \\
  &   &   &  #6 
\end{aligned}
 }%

\title{A Low-cost Mockup to Simulate Robotic Laser Cutting in Nuclear Decommissioning}
\author{Frederico Fernandes Afonso Silva, Murilo Marques Marinho, and Bruno
Vilhena Adorno\thanks{F. F. A. Silva, M. M. Marinho, and B. V. Adorno
are with the Department of Electrical and Electronic Engineering and
the Manchester \foreignlanguage{british}{Centre} for Robotics and
AI, The University of Manchester, United Kingdom (emails: \{frederico.silva;
murilo.marinho; bruno.adorno\}@manchester.ac.uk).}\thanks{This work
was supported by the Royal Academy of Engineering under the Research
Chairs and Senior Research Fellowships \foreignlanguage{british}{programme}
and the Robotics and AI Collaboration (RAICo). }}
\maketitle
\begin{abstract}
This paper introduces a low-cost experimental mockup to simulate the
laser cutting process of containers in nuclear decommissioning. It
is composed of a three-axis table supporting a cuboid container with
ultraviolet-sensitive faces, a six-degree-of-freedom serial manipulator
holding an ultraviolet torch that simulates the laser, and a visual
system based on cameras and fiducial markers. The system employs a
constrained task-space adaptive motion controller that compensates
for inaccurate parameters and eliminates the need to calibrate the
system. Furthermore, as the motion controller explicitly accounts
for geometric constraints, the robot reactively avoids collisions
with obstacles while handling the ultraviolet torch. To improve the
tracking of the laser-cutting path, we control the ultraviolet beam,
which requires only four degrees of freedom, instead of the full end-effector
pose. Experiments show that, despite an initially uncalibrated system,
the overall system is capable of tracking different trajectories with
an overall mean accuracy of $3.9$ (sd $2.5$) mm when the end-effector
pose is controlled and $2.4$ (sd $1.3$) mm when the ultraviolet
beam is controlled.
\end{abstract}

\section{Introduction}

In scenarios where human access is restricted, such as nuclear decommission,
the use of robots can be essential \cite{Jones2026}. Mechanical decontamination
techniques can be used when the contamination is limited to near-surface
material \cite{Laraia2011} and can be automated to reduce human exposure.
However, those strategies require the material to be of specific sizes
and geometries to simplify their processing \cite{Khan2015}.

Robotic laser cutting (RLC) is a suitable strategy for size reduction
of contaminated material \cite{Hilton2014} and minimizes human exposure
to radiation and aerosol hazards during the process \cite{Jones2026}.
However, strict quality standards demand high-precision, which often
require system calibration \cite{Konecny2024,Liu2025,Yang2025}. Given
the varying geometry of the contaminated equipment and the changes
in their topologies as material is removed, the system may require
multiple recalibrations \cite{Hyun2024}.

\begin{figure}
\begin{centering}
\resizebox{1\columnwidth}{!}{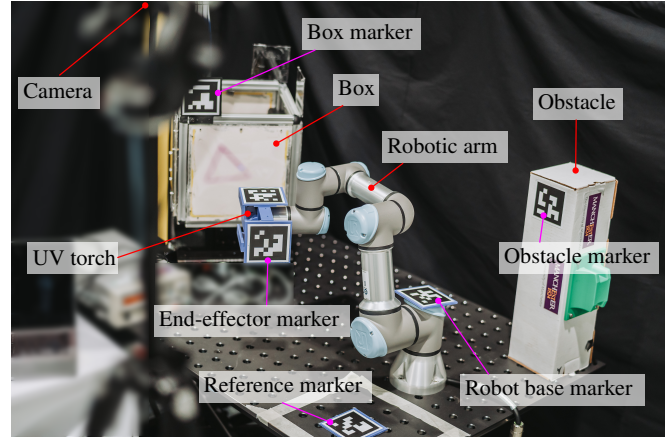}
\par\end{centering}
\caption{Low-cost experimental mockup. An Intel RealSense D435i camera obtains
the poses of fiducial markers attached to each face of a box mounted
on a three-axis table. Markers are attached to obstacles, robot base,
and world frame. In addition, three markers are attached to the robot
end-effector so that at least one marker is visible by the camera
throughout the trajectory. \label{fig:rlc-setup}}
\end{figure}

To deal with this problem, we introduce a low-cost experimental mockup,
shown in Fig.~\ref{fig:rlc-setup}, to simulate RLC of containers
in nuclear decommissioning. It is composed of a three-axis table supporting
a cuboid box with ultraviolet (UV) sensitive faces and a six-degree-of-freedom
(DoF) serial robotic manipulator holding an UV torch that simulates
the laser. The system employs a constrained task-space adaptive motion
controller that compensates for inaccurate parameters and eliminates
the need to calibrate the system. We use a visual system based on
cameras and fiducial markers to provide the adaptive controller with
visual measurements to compensate for the uncalibrated geometric parameters,
and to track obstacles in the workspace. 

\subsection{Related works}

Some RLC solutions in nuclear decommissioning still require intervention
from human operators. For instance, Mallion et al.~\cite{Mallion2017}
deployed a snake-arm robot to dismantle a dissolver vessel in a contaminated
nuclear facility. However, both robot navigation and cutting motion
were manually performed by an operator. Although safer than human
missions inside contaminated facilities \cite{Mallion2016}, remotely
operated robots are still far from the speed efficiency of automated
RLC \cite{Khan2015}.

Addressing this limitation, Ma et al. \cite{Ma2024} proposed a joint-space
path planning algorithm that ensures collision avoidance and satisfy
other application-specific constraints. Nonetheless, motion planning
in joint space has a computational cost that is exponential in the
number of DoFs of the system \cite{LaValle2006book}, and, although
some solutions allow defining constraints on the end-effector pose
\cite{Berenson2011}, defining general task-space constraints can
be quite challenging due to the nonlinear mapping between the task-space
and the joint-space.

Recent works have focused on task-space planning for RLC. For instance,
Anto¨ and Bu¨ek \cite{Antos2024} presented a solution for the serial
production of upholstery products in the automotive industry, which
requires the manual design of a trajectory in a simulation environment,
whereas Hyun et al. \cite{Hyun2024} presented a remote dismantling
system to be deployed in the decommissioning of nuclear power plants
that allows the operator to define high-level task sequences. However,
both solutions require the system to be calibrated in advance to represent
the actual deployment scenario.

To achieve an RLC solution that eliminates the need for system calibration,
enforces safety constraints, is reactive to dynamic changes in the
environment, and can track arbitrary task-space trajectories, we employ
an adaptive constrained task-space controller \cite{Marinho2022}.

\subsection{Statement of contributions}

This paper presents the following contributions:
\begin{enumerate}
\item A complete RLC mockup, shown in Fig.~\ref{fig:rlc-setup}, that employs
UV beams on a photosensitive box mounted on a bespoke $3$-DoF table
for realistic and safe simulation of the RLC process.
\item An extension of the adaptive constrained task-space controller proposed
in \cite{Marinho2022} to control the UV beam instead of the end-effector
pose, which releases two DoFs and increases the robot's functional
redundancy with respect to the laser cutting task. This task relaxation
is done by controlling a line attached to the end-effector that is
always collinear with the UV beam.
\end{enumerate}
We compare the results of following a desired end-effector trajectory
with the results of the relaxed task.Our goal is to verify the system
integration and validate the accuracy of the control strategy for
the intended application. Nonetheless, analysis of laser-cutting phenomena
and related practical requirements of nuclear decommissioning are
outside the scope of this work.

\section{Robotic Laser-cutting Strategy\label{sec:rlc-strategy}}

\begin{figure*}[tph]
\begin{centering}
\fontsize{30}{60}\resizebox{0.95\textwidth}{!}{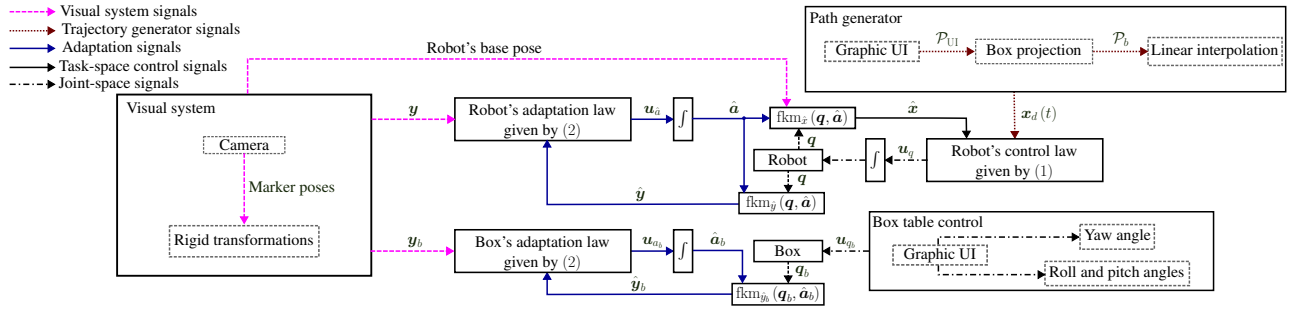}
\par\end{centering}
\caption{Mockup architecture. The visual system provides information for the
adaptive controller to compensate for the system's uncalibrated geometric
parameters while tracking arbitrary, user-defined task-space paths.
Additionally, to simulate skips with different surface orientations,
the box is mounted on top of a $3$-DoF table that can be manually
controlled by the user through a GUI.\label{fig:system-diagram}}
\end{figure*}

The proposed architecture for the low-cost experimental RLC mockup
is shown in Fig.~\ref{fig:system-diagram}. The visual system is
composed of an Intel RealSense D435 camera and seven fiducial markers.
The path generator system allows the user to define a sequence of
points $\mathcal{P}_{\mathrm{UI}}$ containing the endpoints of path
segments through a graphic user interface (GUI). Those endpoints are
then scaled proportionally to the box's face area and projected onto
the box's surface plane, resulting in the sequence $\mathcal{P}_{b}$.
Finally, we perform a linear interpolation between the initial and
final points of each segment in $\mathcal{P}_{b}$ to obtain a desired
cutting path. Additionally, to help us simulate skips (containers)
with different surface orientations, the box is placed on top of a
moving table with three DoFs that can be controlled through a dedicated
GUI.

The adaptive control strategy is divided into two optimization problems
\cite{Marinho2022}. The optimal control inputs, given as joint velocities,
are obtained as
\begin{equation}
\argmintwo q{\left\Vert \mymatrix J_{\hat{x},q}\dot{\myvec q}+\eta_{q}\breve{\myvec x}\right\Vert ^{2}_{2}+\left\Vert \mymatrix{\Lambda}_{q}\dot{\myvec q}\right\Vert ^{2}_{2}}{\mymatrix B_{q}\left(\myvec q,\hat{\myvec a}\right)\dot{\myvec q}\preceq\myvec b_{q}\left(\myvec q,\hat{\myvec a}\right)}{\mymatrix W_{q}\left(\myvec q\right)\dot{\myvec q}\preceq\myvec w_{q}\left(\myvec q\right),}\label{eq:task_optimization_problem}
\end{equation}
where $\dot{\myvec q}\in\mathbb{R}^{n}$ is the configuration-space
velocity vector; $\eta_{q}\in\left(0,\infty\right)$ is a proportional
gain that determines the desired exponential decay rate for the task-space
error, and $\mymatrix{\Lambda}_{q}\in\mathbb{R}^{n\times n}$, with
$\mymatrix{\Lambda}_{q}\succ0$, is a damping matrix; $\mymatrix J_{\hat{x},q}\in\mathbb{R}^{m\times n}$
is the estimated task Jacobian; $\breve{\myvec x}\in\mathbb{R}^{m}$,
with $\breve{\myvec x}=\hat{\myvec x}-\myvec x_{d}$ is the task error,
where $\hat{\myvec x}$ and $\myvec x_{d}$ are the estimated and
desired task vectors, respectively; $\mymatrix B_{q}\in\mathbb{R}^{s\times n}$
and $\myvec b_{q}\in\mathbb{R}^{s}$ define $s$ linear constraints
on the control inputs that depend on the estimated parameters, $\hat{\myvec a}\in\mathbb{R}^{p}$;
and $\mymatrix W_{q}\in\mathbb{R}^{s_{q}\times n}$ and $\myvec w_{q}\in\mathbb{R}^{s_{q}}$
define $s_{q}$ linear constraints on the control inputs that are
unrelated to the estimated parameters \cite{Marinho2022}.

The optimal adaptation signal, that aims to reduce the error $\tilde{\myvec y}\triangleq\hat{\myvec y}-\myvec y=\hat{\myvec x}-\myvec x$
between the estimated task-space vector and the measured task-space
vector is computed as
\begin{equation}
\argminfour{\hat{\myvec a}}{\left\Vert \mymatrix J_{\hat{y},\hat{a}}\dot{\hat{\myvec a}}+\eta_{\hat{a}}\tilde{\myvec y}\right\Vert ^{2}_{2}+\left\Vert \mymatrix{\Lambda}_{\hat{a}}\dot{\hat{\myvec a}}\right\Vert ^{2}_{2}}{\mymatrix B_{\hat{a}}\left(\myvec q,\hat{\myvec a}\right)\dot{\hat{\myvec a}}\preceq\myvec b_{\hat{a}}\left(\myvec q,\hat{\myvec a}\right)}{\mymatrix W_{\hat{a}}\left(\hat{\myvec a}\right)\dot{\hat{\myvec a}}\preceq\myvec w_{\hat{a}}\left(\hat{\myvec a}\right)}{\mymatrix N_{\hat{a}}\dot{\hat{\myvec a}}=\myvec 0}{\breve{\myvec x}^{T}\mymatrix J_{\hat{x},\hat{a}}\dot{\hat{\myvec a}}\leq0,}\label{eq:adaptation_optimization_problem}
\end{equation}
where $\dot{\hat{\myvec a}}\in\mathbb{R}^{p}$ is the time derivative
of the estimated parameters, $\eta_{\hat{a}}\in\left(0,\infty\right)$
is a proportional gain that determines the desired exponential decay
rate for the estimation error, and $\mymatrix{\Lambda}_{\hat{a}}\in\mathbb{R}^{p\times p}$,
with $\mymatrix{\Lambda}_{\hat{a}}\succ0$ is a damping matrix; $\mymatrix J_{\hat{y},\hat{a}}\in\mathbb{R}^{r\times p}$
is the estimated measure-space Jacobian; $\mymatrix B_{\hat{a}}\in\mathbb{R}^{s\times p}$
and $\myvec b_{\hat{a}}\in\mathbb{R}^{s}$ define the linear constraints
that depend on the parameters; $\mymatrix W_{\hat{a}}\in\mathbb{R}^{s_{\hat{a}}\times p}$
and $\myvec w_{\hat{a}}\in\mathbb{R}^{s_{\hat{a}}}$ define $s_{\hat{a}}$
linear constraints that are independent of the robot configuration;
$\mymatrix N_{\hat{a}}\in\mathbb{R}^{s_{\hat{a}}\times p}$ is the
parametric task-Jacobian projector that prevents disturbing unmeasured
variables; and $\mymatrix J_{\hat{x},\hat{a}}\in\mathbb{R}^{m\times p}$
is the estimated parametric Jacobian \cite{Marinho2022}.\footnote{For a general analysis of the effects of initial calibration errors
and parameter uncertainties on the convergence of the estimated model,
please refer to \cite{Marinho2022}.}

The adaptive task-space control laws given by (\ref{eq:task_optimization_problem})
and (\ref{eq:adaptation_optimization_problem}) are very general and
require a model given in task-space representation. In the present
work, the task-space vector can be either a parametrized version of
the end-effector pose or the coefficients of a line that is collinear
to the UV beam \cite{Adorno2017book}. Nonetheless, any parametrization
can be used, as long as the corresponding Jacobian is available. Additionally,
the task error is directly calculated in the task space, and robot
motion control of the robot does not require path planning.

We use this adaptive control strategy for both the robot and the box
to compensate for the uncalibrated geometric parameters across the
system. More specifically, the relative pose between the reference
marker and the robot base frame is uncertain, as well as the mechanical
attachment of the UV torch on the end-effector. Furthermore, although
the relative pose between the camera and the fiducial markers attached
to the skip can be directly measured, the attachment location is inaccurate.
Another source of uncertainty is the actual placement of the skip
on the bespoke $3$-DoF moving table, which can be modeled as a spherical
joint.

\subsection{Visual system}

We use a visual system based on a Intel RealSense D435 camera and
fiducial markers\footnote{\textcolor{black}{We resorted to fiducial markers in the mockup design
for the sake of simplicity, but their use would be unpractical in
actual nuclear decommissioning. We intend to address this limitation
in future works.}} to obtain the pose information needed for the feedback loop in the
adaptive law (\ref{eq:adaptation_optimization_problem}) and for the
definition of safety constraints. Namely, we obtain the poses of each
fiducial marker from the camera's API and calculate the nominal rigid
transformations to retrieve the robot base's and end-effector's poses,
as well as the poses of the box representing the container and the
poses of any possible obstacles in the environment, all with respect
to the reference frame. These rigid transformations are inaccurate
because the mechanical attachments of the fiducial markers are imperfect
and contain small placement errors.

Since the end-effector can be partially occluded during the robot's
motion, we used a 3-D printed device to hold three different markers
around the UV torch. This way, the visual system can use any of them
to measure the end-effector pose.

\subsection{Safety constraints\label{subsec:Safety-constraints}}

We use vector field inequalities (VFIs) \cite{Marinho2019} to enforce
hard constraints based on the geometry of the workspace. Each pair
of geometric primitives has an associated VFI that ensure they never
intersect, guaranteeing safety. Furthermore, each VFI contributes
to one row in $\mymatrix B_{q},\mymatrix W_{q},\mymatrix B_{\hat{a}},\mymatrix W_{\hat{a}}$
and one coefficient in $\myvec b_{q},\myvec w_{q},\myvec b_{\hat{a}},\myvec w_{\hat{a}}$.
In general, given the signed distance $\mathsf{d}\left(\myvec g_{r}\left(\myvec q,\hat{\quat a}\right),\myvec g_{w}\right)\in\mathbb{R}$
between a geometric primitive $\myvec g_{r}(\myvec q,\hat{\myvec a})$
attached to the robot and a geometric primitive $\myvec g_{w}$ in
the workspace, each VFI in (\ref{eq:task_optimization_problem}) is
defined as 
\begin{equation}
\underset{i\text{th row in }\mymatrix B_{q}}{\underbrace{-\frac{\partial\mathsf{d}\left(\myvec g_{r}\left(\myvec q,\hat{\quat a}\right),\myvec g_{w}\right)}{\partial\myvec q}}}\dot{\myvec q}\leq\underset{i\text{th coefficient in }\myvec b_{q}}{\underbrace{\eta_{\mathrm{vfi},q}\mathsf{d}\left(\myvec g_{r}\left(\myvec q,\hat{\quat a}\right),\myvec g_{w}\right)}}\label{eq:VFI-outside}
\end{equation}
when the goal is to keep $\myvec g_{r}\left(\myvec q,\hat{\quat a}\right)$
\emph{outside} of $\myvec g_{w}$. Conversely, if the goal is to keep
$\myvec g_{r}\left(\myvec q,\hat{\quat a}\right)$ \emph{inside} $\myvec g_{w}$,
the corresponding VFI is defined as 
\begin{equation}
\frac{\partial\mathsf{d}\left(\myvec g_{r}\left(\myvec q,\hat{\quat a}\right),\myvec g_{w}\right)}{\partial\myvec q}\dot{\myvec q}\leq-\eta_{\mathrm{vfi},q}\mathsf{d}\left(\myvec g_{r}\left(\myvec q,\hat{\quat a}\right),\myvec g_{w}\right).\label{eq:VFI-inside}
\end{equation}
Analogous VFIs with respect to $\hat{\myvec a}$ are used in (\ref{eq:adaptation_optimization_problem}).
More details can be found in \cite{Marinho2022}.

\begin{figure}
\begin{centering}
\resizebox{0.9\columnwidth}{!}{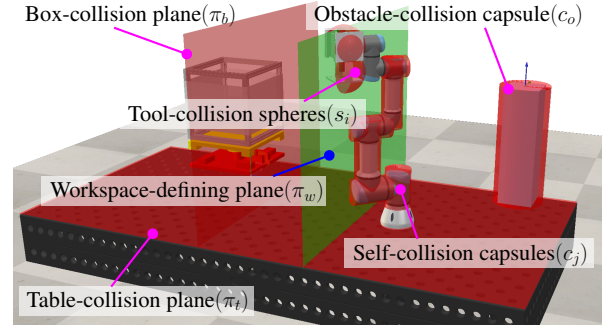}
\par\end{centering}
\caption{Digital-twin. Geometric primitives in red indicate zones for the robot's
end-effector to avoid, whereas the green plane indicates the beginning
of the operating zone for the end-effector.\label{fig:Digital-twin}}
\end{figure}
The end-effector is tightly enclosed with two spheres $s_{i}$, $i\in\left\{ 1,2\right\} $
with a radius of 0.075~m, whereas the robot's links are tightly enclosed
with six capsules $c_{j}$, $j\in\left\{ 1,2,3,4,5,6\right\} $. The
first three capsules have a radius of 0.005~m, the fourth has a radius
of 0.045~m, and the last two have a radius of 0.04~m, as shown in
Fig.~\ref{fig:Digital-twin}. Additionally, the obstacle is enclosed
with a capsule $c_{o}$ with a radius of 0.1~m and the planes $\pi_{t}$
and $\pi_{b}$ are used to prevent collisions with the table that
supports the robotic manipulator and the box, respectively. Finally,
the plane $\pi_{w}$ is used to define the beginning of the operating
zone for the robot because the distance between the laser source and
the cutting surface affects the cutting quality. As such, there are
a total of 19 VFIs, namely: four VFIs to prevent the collision between
the two spheres $s_{1},s_{2}$ and the two planes $\pi_{t},\pi_{b}$;
12 VFIs for the pair $\left(s_{i},c_{j}\right)$, for all $i,j$,
to prevent self-collisions; two VFIs for the pair $\left(s_{i},c_{o}\right)$,
for all $i$, to prevent collisions between the end-effector and the
obstacle; and one VFI between $s_{1}$ and $\pi_{w}$ to ensure the
end-effector stays within the operating area.

Analogously, there are 19 VFIs associated with the same geometric
primitives for the 36 uncertain parameters, namely: all Denavit-Hartenberg
parameters (24 in total) and an initial rough estimate of the base
frame $\frame b$ (six) and end-effector frame $\frame{\mathrm{eff}}$
(six). The transformations $\frame b$ and $\frame{\mathrm{eff}}$
represent three translations along and three rotations around the
$x$-axis, $y$-axis, and $z$-axis, in this respective order. The
initial estimations were obtained using a metric tape.

Finally, the joint velocities are limited by defining $\mymatrix W_{q}=\left[\begin{array}{cc}
-\mymatrix I_{6\times6} & \mymatrix I_{6\times6}\end{array}\right]^{T}$ and $\myvec b_{q}=\left[\begin{array}{cc}
-\myvec q^{T}_{v,\mathrel{\min}} & \myvec q^{T}_{v,\mathrel{\max}}\end{array}\right]^{T}$.

\subsection{Task relaxation to control the ultraviolet beam: the line control
objective\label{subsec:task-relaxation}}

Beams, e.g. laser or UV, have axial symmetry so rotations around the
beam axis does not affect the resulting projected path in the surface.
Therefore, controlling the end-effector pose becomes unnecessary.
A better approach is to control a line that is collinear with the
beam so that the rotation around and translation about its axis is
not controlled, releasing two DoF and creating a functional redundancy
with respect to the task. The control objective is then to make this
line converge to the desired lines along the cutting path. For details
on how to obtain the line parameters, the corresponding line Jacobian,
and how to calculate the induced distance between lines, refer to
\cite{Marinho2019}.

\section{Experiments\label{sec:Experiments}}

We assessed the closed-loop control strategy illustrated in Fig.~\ref{fig:system-diagram}.
This way, all the pipeline, including the perception system, path
generator, and the adaptive constrained task-space controller can
be evaluated in terms of the system's ability to track a desired laser
cutting path. We used the computational library DQ Robotics \cite{Adorno2021}
to model the robot, the geometric constraints, and to implement the
control law in C++. We used the SmartArmStack framework \cite{Marinho2024}
for communication with the robot and CoppeliaSim Edu V4.7.0 \cite{Rohmer2013}
as a digital-twin for visualization during the experiments, as shown
in Fig.~\ref{fig:Digital-twin}. The parameters used for the controllers
are given in Table~\ref{tab:experiment-parameters}.

\begin{figure}
\begin{centering}
\fontsize{130}{60}\resizebox{0.7\columnwidth}{!}{
\begingroup%
  \makeatletter%
  \providecommand\color[2][]{%
    \errmessage{(Inkscape) Color is used for the text in Inkscape, but the package 'color.sty' is not loaded}%
    \renewcommand\color[2][]{}%
  }%
  \providecommand\transparent[1]{%
    \errmessage{(Inkscape) Transparency is used (non-zero) for the text in Inkscape, but the package 'transparent.sty' is not loaded}%
    \renewcommand\transparent[1]{}%
  }%
  \providecommand\rotatebox[2]{#2}%
  \newcommand*\fsize{\dimexpr\f@size pt\relax}%
  \newcommand*\lineheight[1]{\fontsize{\fsize}{#1\fsize}\selectfont}%
  \ifx\svgwidth\undefined%
    \setlength{\unitlength}{3468.00004614bp}%
    \ifx\svgscale\undefined%
      \relax%
    \else%
      \setlength{\unitlength}{\unitlength * \real{\svgscale}}%
    \fi%
  \else%
    \setlength{\unitlength}{\svgwidth}%
  \fi%
  \global\let\svgwidth\undefined%
  \global\let\svgscale\undefined%
  \makeatother%
  \begin{picture}(1,0.75086507)%
    \lineheight{1}%
    \setlength\tabcolsep{0pt}%
    \put(0,0){\includegraphics[width=\unitlength,page=1]{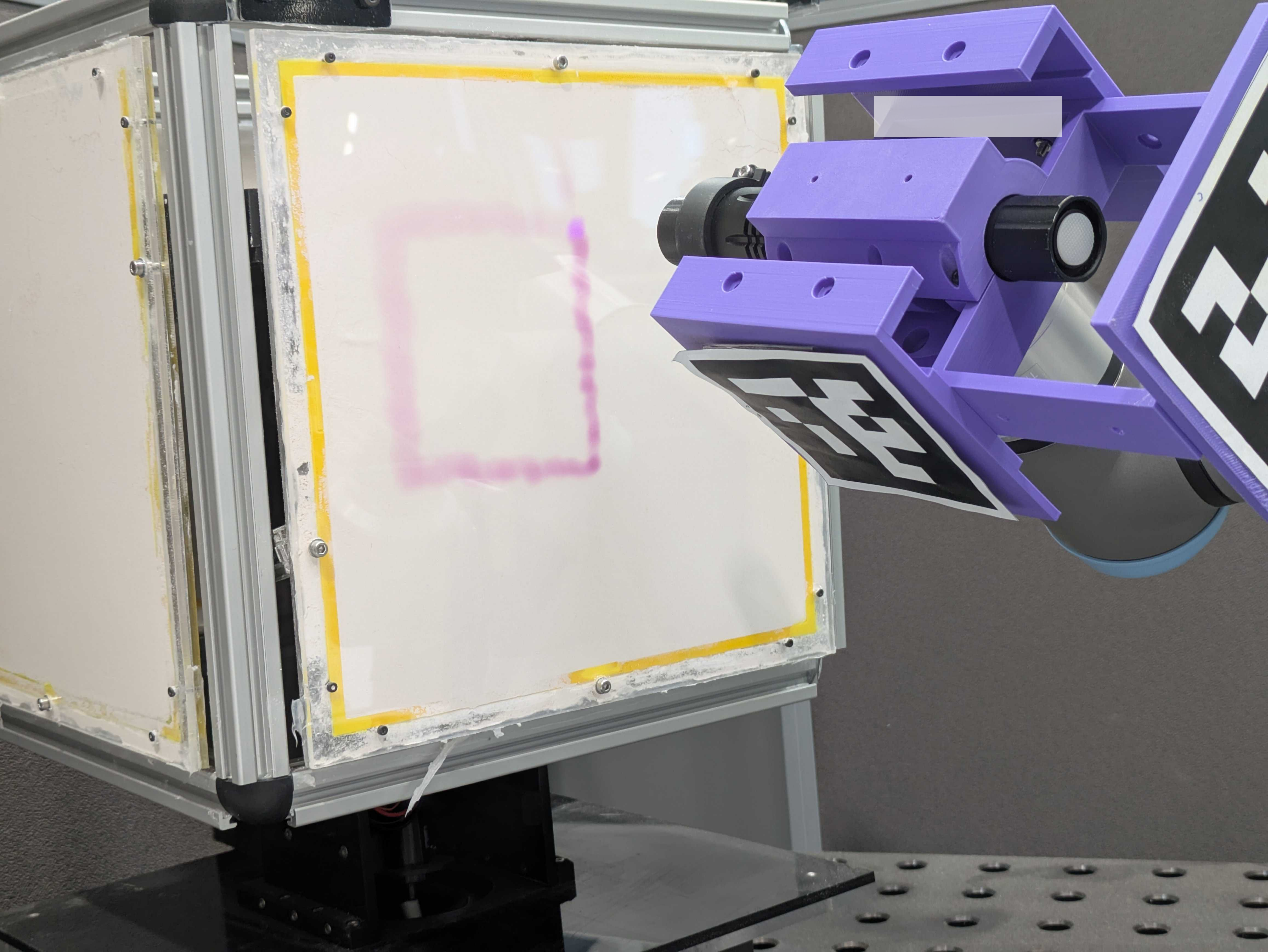}}%
    \put(0.69228473,0.64763866){\color[rgb]{0,0,0}\makebox(0,0)[lt]{\lineheight{1.25}\smash{\begin{tabular}[t]{l}UV  torch\end{tabular}}}}%
    \put(0,0){\includegraphics[width=\unitlength,page=2]{images/box_drawing.pdf}}%
    \put(0.53236996,0.24084879){\color[rgb]{0,0,0}\makebox(0,0)[lt]{\lineheight{1.25}\smash{\begin{tabular}[t]{l}Reaction to UV light exposure\end{tabular}}}}%
    \put(0,0){\includegraphics[width=\unitlength,page=3]{images/box_drawing.pdf}}%
    \put(0.01513158,0.70003275){\color[rgb]{0,0,0}\makebox(0,0)[lt]{\lineheight{1.25}\smash{\begin{tabular}[t]{l}Photochromic powder\end{tabular}}}}%
    \put(0,0){\includegraphics[width=\unitlength,page=4]{images/box_drawing.pdf}}%
    \put(0.70508538,0.02252966){\color[rgb]{0,0,0}\makebox(0,0)[lt]{\lineheight{1.25}\smash{\begin{tabular}[t]{l}$3$-DoF table\end{tabular}}}}%
    \put(0,0){\includegraphics[width=\unitlength,page=5]{images/box_drawing.pdf}}%
  \end{picture}%
\endgroup%
}
\par\end{centering}
\caption{Cuboid container with UV sensitive faces. The acrylic surfaces are
filled with a photochromic powder that changes colors when exposed
to UV light.\label{fig:box-drawing}}
\end{figure}

We used a $6$-DoF UR3e robotic manipulator equipped with a RS PRO
LED UV torch and a cuboid container to simulate the laser cutting
process, as shown in Fig.~\ref{fig:box-drawing}. The container has
acrylic surfaces filled with a photochromic pigment powder that changes
color when exposed to UV light, and is mounted on top of a $3$-DoF
bespoke table. Nonetheless, as we use a parametric representation
of the robot, the robotic manipulator can be easily replaced as suitable
for deployment.

We actuate the table only in the beginning of the experiment using
a GUI in order to rotate the container to simulate the different surfaces
orientations that would be present when laser-cutting arbitrary skips.
Then, before the robot starts moving, the adaptive control law (\ref{eq:adaptation_optimization_problem})
is run for an arbitrary period of $5$s to improve the initial estimated
model, which might be very inaccurate due to the lack of calibration.
Subsequently, all VFI constraints described in Section~\ref{subsec:Safety-constraints}
are added to the constrained controller, except for the constraint
related to the plane $\pi_{w}$, so that the robot end-effector can
move to a predefined pose between planes $\pi_{w}$ and $\pi_{t}$.
Finally, once the end-effector is within the operating area, the remaining
VFI related to $\pi_{w}$ is added to the optimization problem and
control laws (\ref{eq:task_optimization_problem}) and (\ref{eq:adaptation_optimization_problem})
are run continuously in a closed-loop so that the desired path is
tracked while the robot adapts its uncertain parameters continuously.

\begin{table}
\caption{Parameters for the control law (\ref{eq:task_optimization_problem})
and adaptation law.\label{tab:experiment-parameters}}

\begingroup
\setlength{\tabcolsep}{3pt} 
\begin{centering}
\begin{tabular*}{1\columnwidth}{@{\extracolsep{\fill}}cccccc}
\hline 
$\eta_{q}$ & $\eta_{\mathrm{vfi},q},\eta_{\hat{a}},\eta_{\mathrm{vfi},\hat{a}}$ & $\mymatrix{\Lambda}_{q}$ & $\mymatrix{\Lambda}_{\hat{a}}$ & $\myvec q_{v,\mathrel{\min}}\left(\unitfrac{rad}{s}\right)$ & $\myvec q_{v,\mathrel{\max}}\left(\unitfrac{rad}{s}\right)$\tabularnewline
\hline 
$50$ & $5$ & $0.02\mymatrix I_{6}$ & $0.02\mymatrix I_{36}$ & $-0.2\cdot\mymatrix 1_{6}$ & $0.2\cdot\mymatrix 1_{6}$\tabularnewline
\hline 
\end{tabular*}
\par\end{centering}
\endgroup
\end{table}

\subsection{Definition of cutting paths \label{subsec:cutting-path-definition}}

Using the procedure described in Section~\ref{sec:rlc-strategy},
we defined four arbitrary cutting paths: a vertical line, a square,
a triangle, and a diamond. Those cutting paths were used to generate
pose paths for the adaptive controller with pose control objective
(ACPO) and\emph{ }line paths for the adaptive controller with line
control objective (ACLO).

The pose paths for ACPO are generated as follows. For each point along
the cutting path, the end-effector orientation is defined such that
its $z$-axis is normal and opposite to the box's surface. The end-effector
position is defined so that the UV torch is $15$~cm away from the
box. For each initial and final point of a given segment in the cutting
path, we interpolate the pose path using screw linear interpolation
\cite{Sarker2020}.

\begin{figure}
\begin{centering}
\resizebox{0.5\columnwidth}{!}{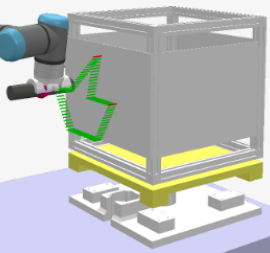}
\par\end{centering}
\caption{Line paths for ACLO. Reference lines perpendicular to the box surface
are assigned to each point along the interpolated cutting path.\label{fig:ACLO-trajectory-generation}}
\end{figure}

The line paths for ACLO are defined by assigning a reference line
that is perpendicular to the estimated box's surface for each point
along the interpolated cutting path, as illustrated in Fig.~\ref{fig:ACLO-trajectory-generation}.

\subsection{Analysis of the pose control objective\label{subsec:Analysis-of-acpo}}

When using the ACPO to track a desired path, there are an infinite
number of orientations that keep the UV torch perpendicular to the
box. This is due to the fact that the torch can be continuously rotated
around its axis without changing the beam direction. However, when
enforcing a pre-defined end-effector orientation, the robot might
reach configurations near singularities in order to attain the desired
end-effector pose while respecting all constraints, which can potentially
lead to unstable behavior.

\begin{figure}
\begin{centering}
\subfloat[Measured positions of the end-effector when using ACPO to track $\protect\myvec x_{d_{n}}\left(\lambda\right)$
(\emph{blue} curve), which is a pose path with carefully chosen orientations,
and $\protect\myvec x_{d_{m}}\left(\lambda\right)$ (\emph{red} curve),
which is a path with the same positions as $\protect\myvec x_{d_{n}}\left(\lambda\right)$
but with a $30^{\circ}$ rotation angle around the end-effector $z$-axis.
The \emph{solid black} line represents the desired end-effector positions.\label{fig:ACPO-square-trajectory-unstable}]{\begin{centering}
\fontsize{20}{60}%
\begin{minipage}[t]{0.95\columnwidth}%
\begin{center}
\resizebox{0.8\columnwidth}{!}{
\begingroup%
  \makeatletter%
  \providecommand\color[2][]{%
    \errmessage{(Inkscape) Color is used for the text in Inkscape, but the package 'color.sty' is not loaded}%
    \renewcommand\color[2][]{}%
  }%
  \providecommand\transparent[1]{%
    \errmessage{(Inkscape) Transparency is used (non-zero) for the text in Inkscape, but the package 'transparent.sty' is not loaded}%
    \renewcommand\transparent[1]{}%
  }%
  \providecommand\rotatebox[2]{#2}%
  \newcommand*\fsize{\dimexpr\f@size pt\relax}%
  \newcommand*\lineheight[1]{\fontsize{\fsize}{#1\fsize}\selectfont}%
  \ifx\svgwidth\undefined%
    \setlength{\unitlength}{598.38368225bp}%
    \ifx\svgscale\undefined%
      \relax%
    \else%
      \setlength{\unitlength}{\unitlength * \real{\svgscale}}%
    \fi%
  \else%
    \setlength{\unitlength}{\svgwidth}%
  \fi%
  \global\let\svgwidth\undefined%
  \global\let\svgscale\undefined%
  \makeatother%
  \begin{picture}(1,0.59294034)%
    \lineheight{1}%
    \setlength\tabcolsep{0pt}%
    \put(0,0){\includegraphics[width=\unitlength,page=1]{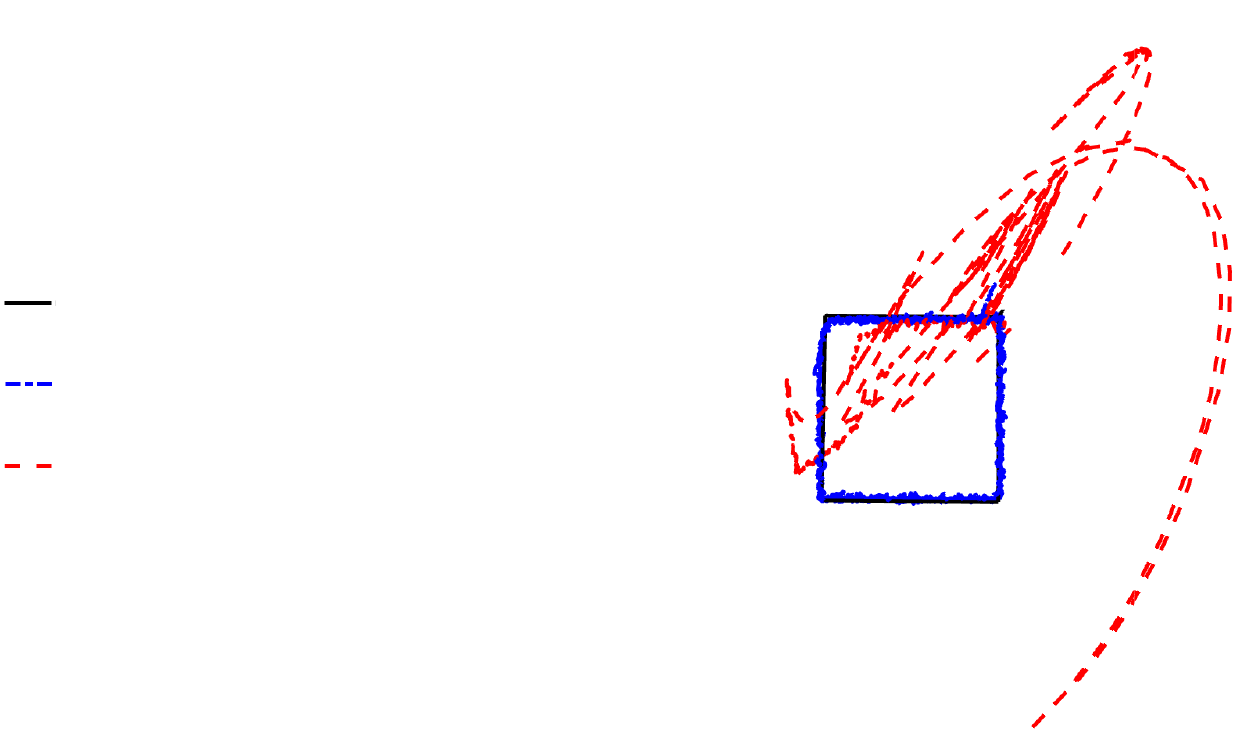}}%
    \put(0.05240071,0.33829458){\makebox(0,0)[lt]{\lineheight{1.25}\smash{\begin{tabular}[t]{l}Desired end-effector position\end{tabular}}}}%
    \put(0.05240071,0.2116891){\makebox(0,0)[lt]{\lineheight{1.25}\smash{\begin{tabular}[t]{l}ACPO: unstable closed-loop system\end{tabular}}}}%
    \put(0.05240071,0.27784021){\makebox(0,0)[lt]{\lineheight{1.25}\smash{\begin{tabular}[t]{l}ACPO: stable closed-loop system\end{tabular}}}}%
  \end{picture}%
\endgroup%
}
\par\end{center}%
\end{minipage}
\par\end{centering}
}
\par\end{centering}
\begin{centering}
\subfloat[Closed-loop error response and smallest non-zero singular value. The
norm of the error is given by the dashed black line, whereas the magenta
line shows the smallest non-zero singular value of the task Jacobian.\label{fig:acpo-norm-error-singularity}]{\begin{centering}
\fontsize{40}{60}%
\begin{minipage}[t]{0.95\columnwidth}%
\begin{center}
\resizebox{0.8\columnwidth}{!}{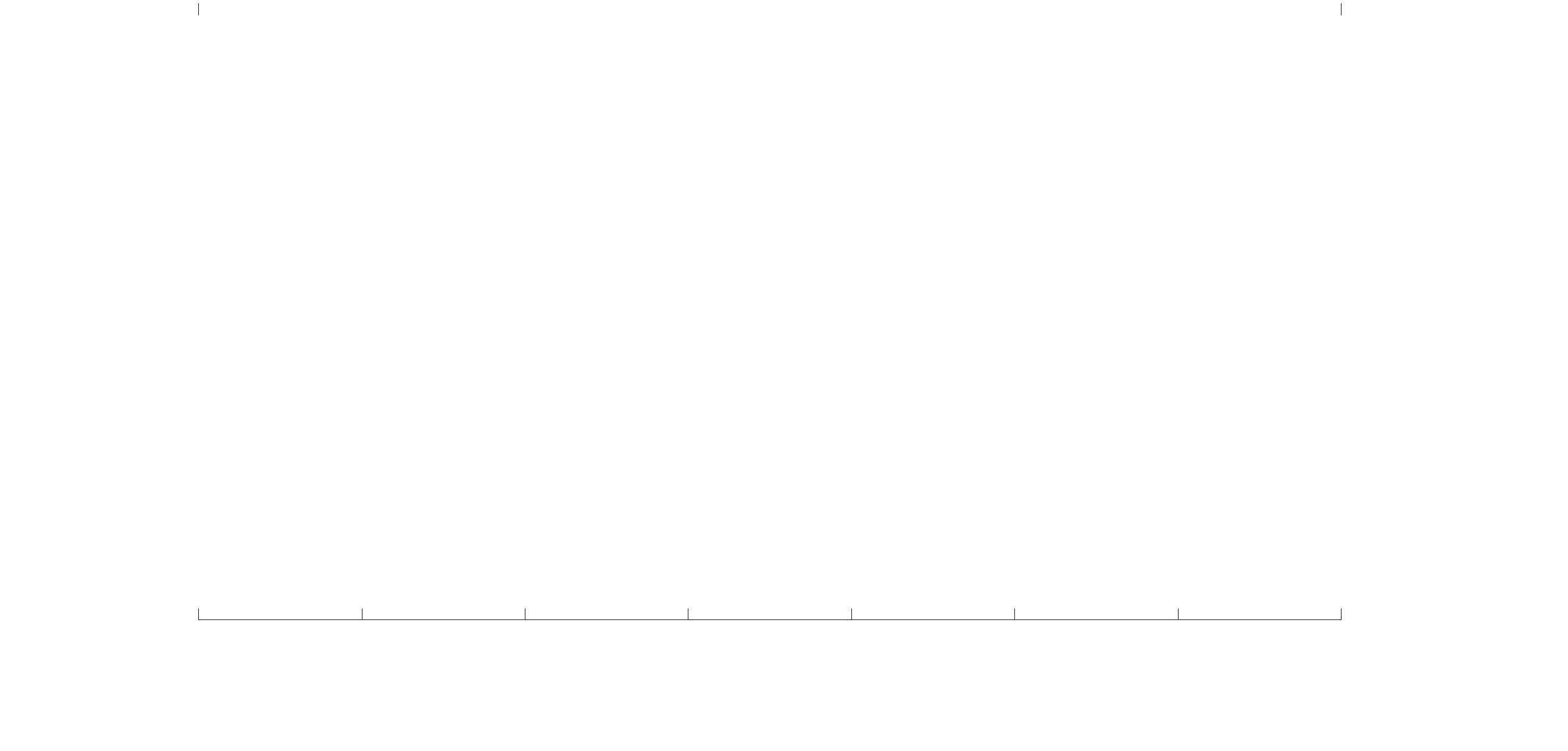}
\par\end{center}%
\end{minipage}
\par\end{centering}
}
\par\end{centering}
\caption{Effects of different rotation angles around the UV beam axis, which
is collinear with the end-effector $z$-axis, when using ACPO to follow
paths with the same desired end-effector positions.\label{fig:ACPO-varying-orientation}}
\end{figure}

For instance, Fig.~\ref{fig:ACPO-square-trajectory-unstable} shows
the UR3e following two desired pose paths: (1) a nominal pose path
$\myvec x_{d_{n}}\left(\lambda\right)$, with executed path in \emph{blue},
representing a square with the end-effector's $z$-axis perpendicular
to the surface and the $x$- and $y$-axes carefully chosen so that
the task Jacobian would be well-conditioned at all times; (2) and
a modified pose path $\myvec x_{d_{m}}\left(\lambda\right)$, with
executed path in \emph{red}, in which the end-effector was rotated
by $30{^\circ}$ about the torch's axis with respect to the nominal
desired path. Whereas the closed-loop under ACPO is well behaved when
tracking the carefully defined $\myvec x_{d_{n}}\left(\lambda\right)$,
it tends to become unstable when tracking $\myvec x_{d_{m}}\left(\lambda\right)$.
This can be explained by inspecting Fig.~\ref{fig:acpo-norm-error-singularity},
that shows the error norm decay and the smallest non-zero singular
value of the task Jacobian when tracking $\myvec x_{d_{m}}\left(\lambda\right)$,
which provides a measurement of the robot manipulability. When the
smallest singular value decreases substantially or abruptly, the manipulability
is affected and the error norm increases substantially. This phenomenon
highlights the importance of not relying on a pre-selection of the
end-effector orientation for the beam control, which can be achieved
by relaxing the task as discussed in Section~\ref{subsec:task-relaxation}.
Fig.~\ref{fig:ACPO-square-trajectory-unstable} shows the measured
end-effector positions, which are retrieved from the fiducial markers
by the visual system. When close to singularities, the abrupt pose
variations introduce noise spikes into the measurements. This leads
to the artificial amplification of the measured end-effector's movement\footnote{A more detailed analysis of the effects of marker noise is outside
the scope of this work. Future work will focus on a more robust visual
system.} shown in Fig.~\ref{fig:ACPO-square-trajectory-unstable}.

Thus, for a fair comparison between ACPO and ACLO in Section~\ref{subsec:Comparison-acpo-aclo},
we selected desired pose paths for the ACPO that do not take the robot
configuration close to singularities.

\subsection{Comparison of control objectives\label{subsec:Comparison-acpo-aclo}}

\begin{figure}
\begin{centering}
\subfloat[Vertical line path.\label{fig:line-trajectory}]{\begin{centering}
\fontsize{32}{60}\resizebox{0.3\columnwidth}{!}{
\begingroup%
  \makeatletter%
  \providecommand\color[2][]{%
    \errmessage{(Inkscape) Color is used for the text in Inkscape, but the package 'color.sty' is not loaded}%
    \renewcommand\color[2][]{}%
  }%
  \providecommand\transparent[1]{%
    \errmessage{(Inkscape) Transparency is used (non-zero) for the text in Inkscape, but the package 'transparent.sty' is not loaded}%
    \renewcommand\transparent[1]{}%
  }%
  \providecommand\rotatebox[2]{#2}%
  \newcommand*\fsize{\dimexpr\f@size pt\relax}%
  \newcommand*\lineheight[1]{\fontsize{\fsize}{#1\fsize}\selectfont}%
  \ifx\svgwidth\undefined%
    \setlength{\unitlength}{576.63478088bp}%
    \ifx\svgscale\undefined%
      \relax%
    \else%
      \setlength{\unitlength}{\unitlength * \real{\svgscale}}%
    \fi%
  \else%
    \setlength{\unitlength}{\svgwidth}%
  \fi%
  \global\let\svgwidth\undefined%
  \global\let\svgscale\undefined%
  \makeatother%
  \begin{picture}(1,1.11295253)%
    \lineheight{1}%
    \setlength\tabcolsep{0pt}%
    \put(0,0){\includegraphics[width=\unitlength,page=1]{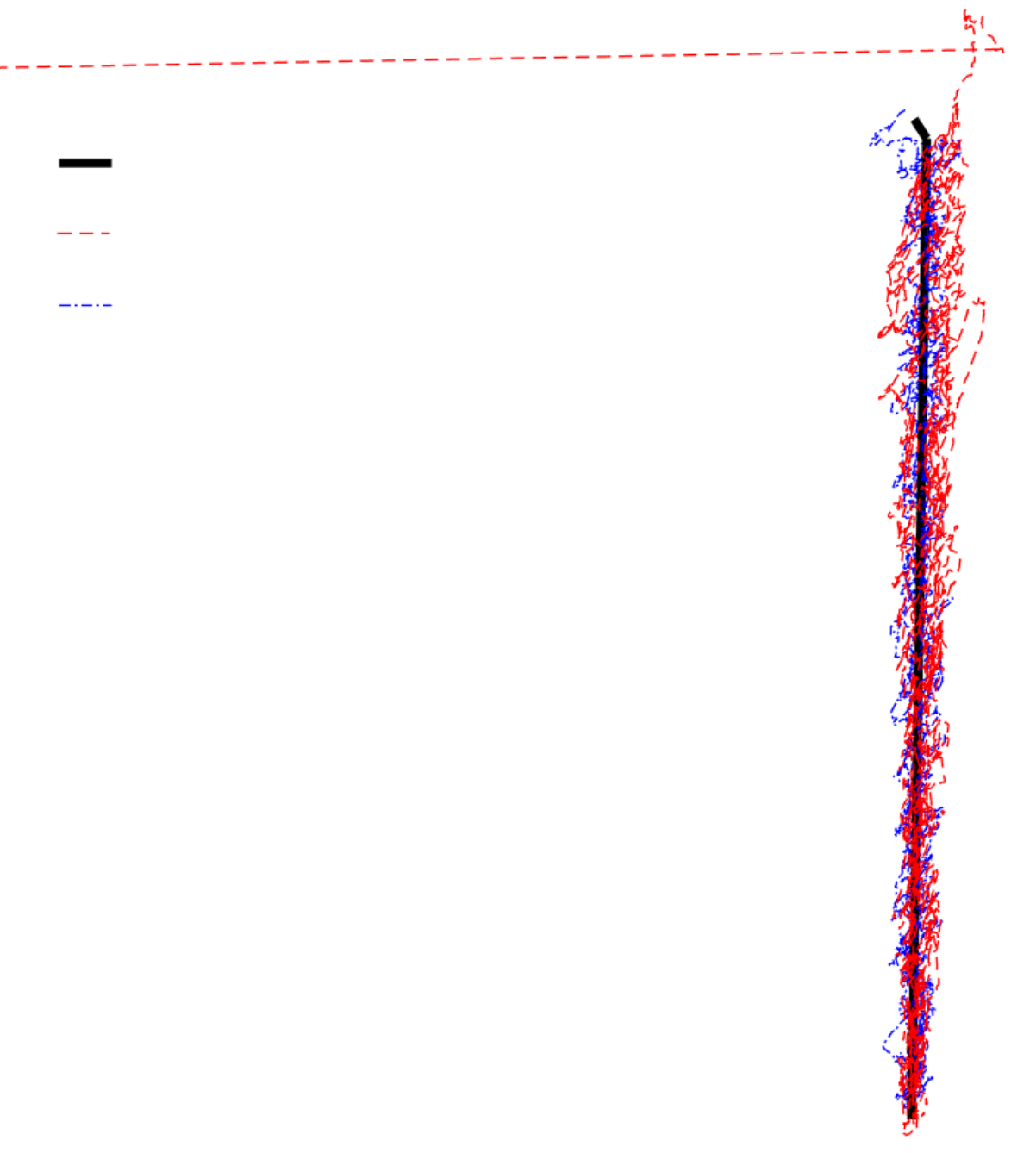}}%
    \put(0.12085887,0.94150284){\makebox(0,0)[lt]{\lineheight{1.25}\smash{\begin{tabular}[t]{l}Desired end-effector position\end{tabular}}}}%
    \put(0.12085887,0.87586217){\makebox(0,0)[lt]{\lineheight{1.25}\smash{\begin{tabular}[t]{l}ACPO real end-effector position\end{tabular}}}}%
    \put(0.12085887,0.80721605){\makebox(0,0)[lt]{\lineheight{1.25}\smash{\begin{tabular}[t]{l}ACLO real end-effector position\end{tabular}}}}%
    \put(0,0){\includegraphics[width=\unitlength,page=2]{images/end_effector_trajectory/line_trajectory.pdf}}%
  \end{picture}%
\endgroup%
}
\par\end{centering}
}~\subfloat[Square path.\label{fig:square-trajectory}]{\begin{centering}
\resizebox{0.3\columnwidth}{!}{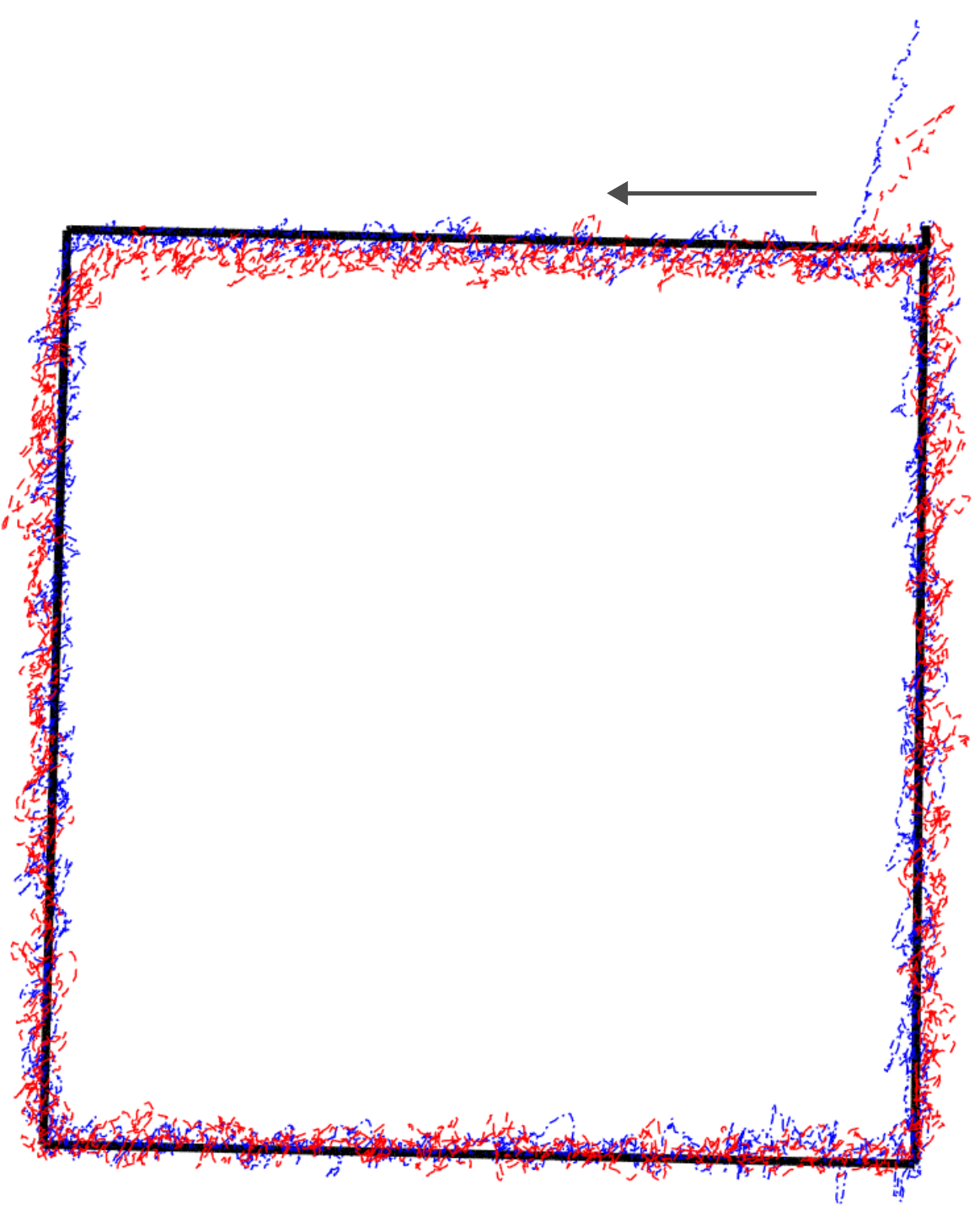}
\par\end{centering}
}
\par\end{centering}
\begin{centering}
\subfloat[Triangle path.\label{fig:triangle-trajectory}]{\begin{centering}
\resizebox{0.4\columnwidth}{!}{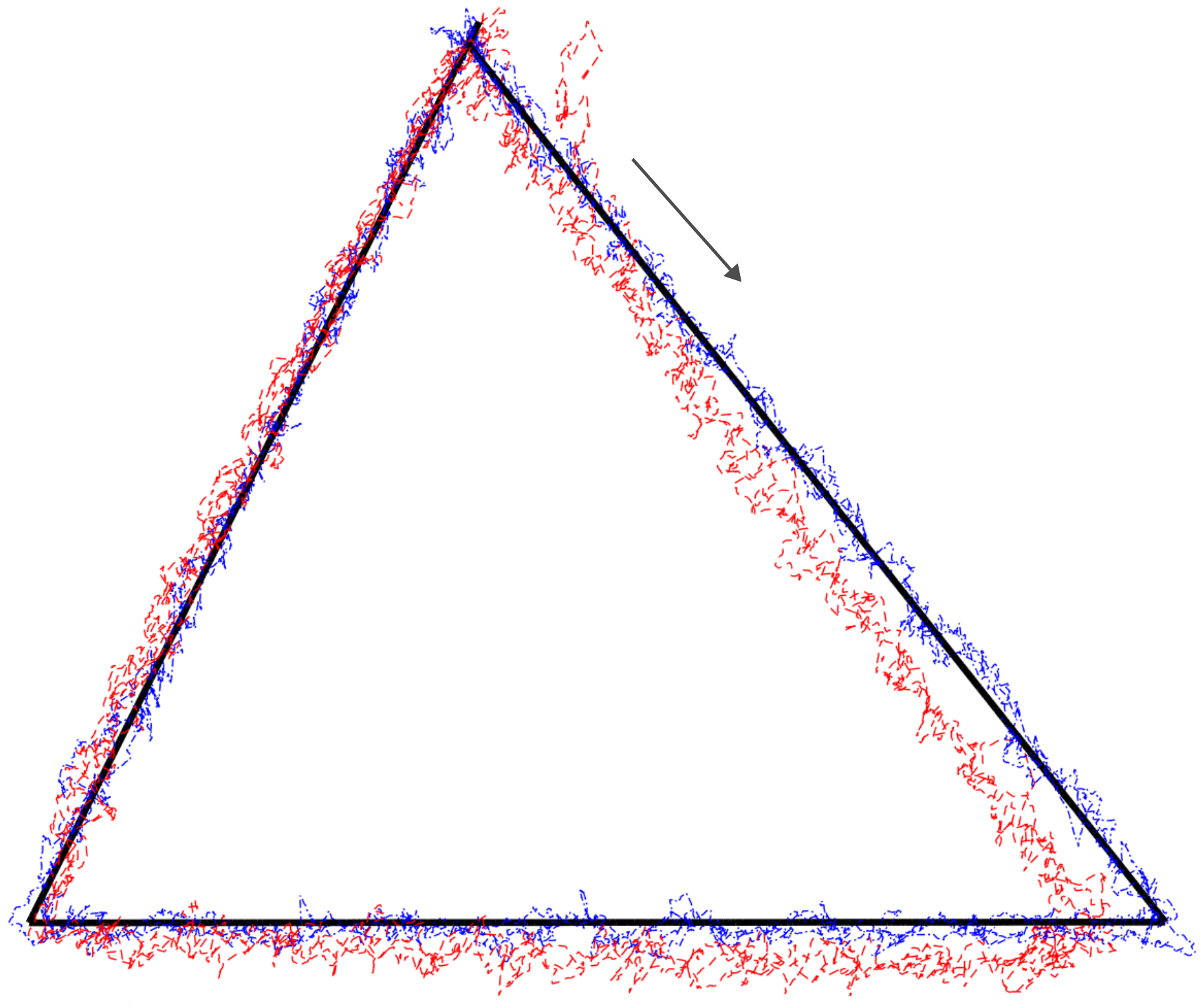}
\par\end{centering}
}~\subfloat[Diamond path.\label{fig:diamond-trajectory}]{\begin{centering}
\resizebox{0.3\columnwidth}{!}{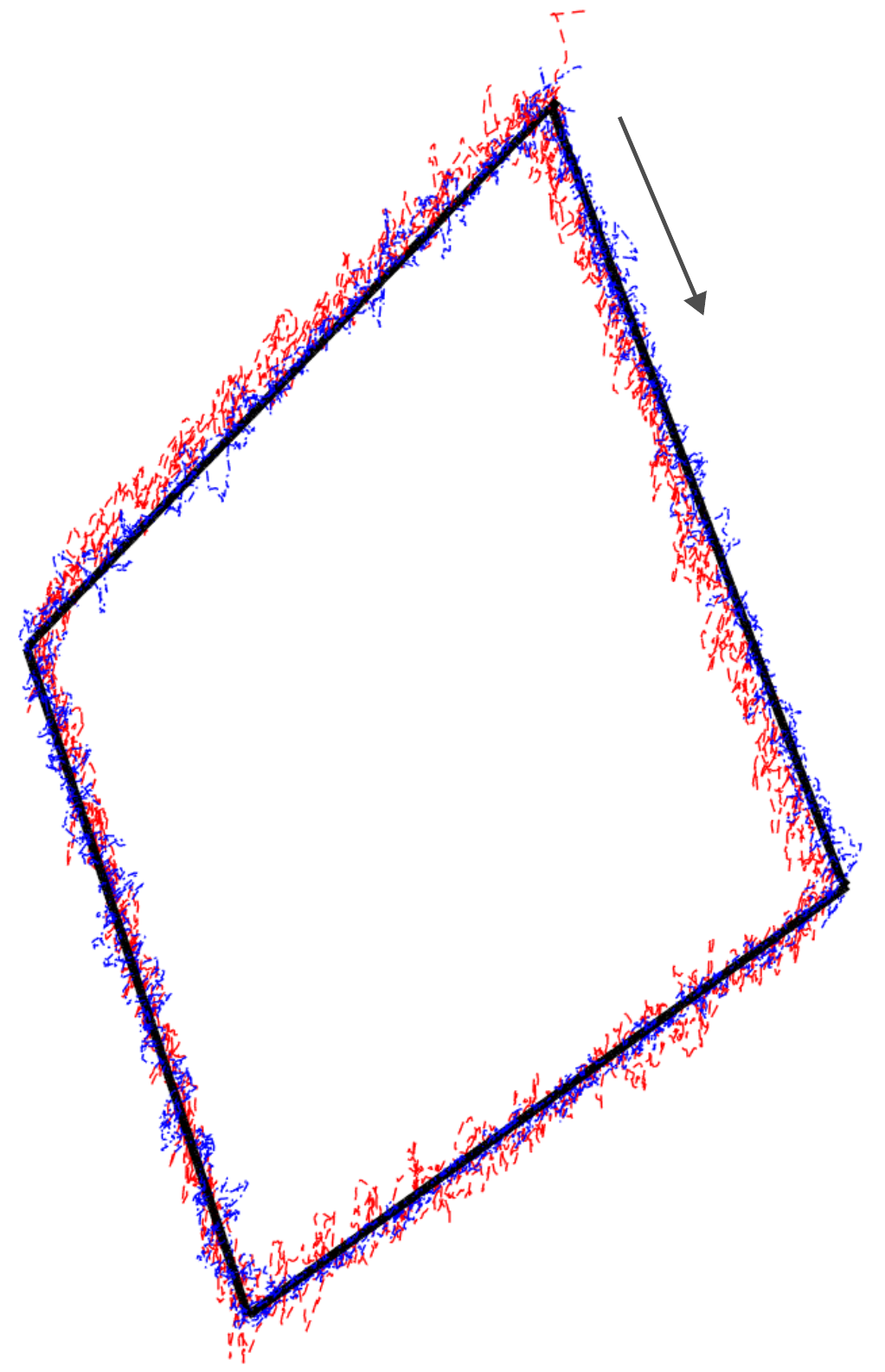}
\par\end{centering}
}
\par\end{centering}
\caption{Traced path while tracking different paths over three circulations.
The solid black line represents the desired path, whereas the red
dashed line indicates the traced path when using ACPO, and the blue
dot-dashed line shows the traced path when using ACLO. The gray arrows
indicate the circulation direction. \label{fig:end-effector-trajectories}}
\end{figure}

\begin{figure}
\begin{centering}
\subfloat[Vertical line.]{\begin{centering}
\fontsize{25}{60}\resizebox{0.4\textwidth}{!}{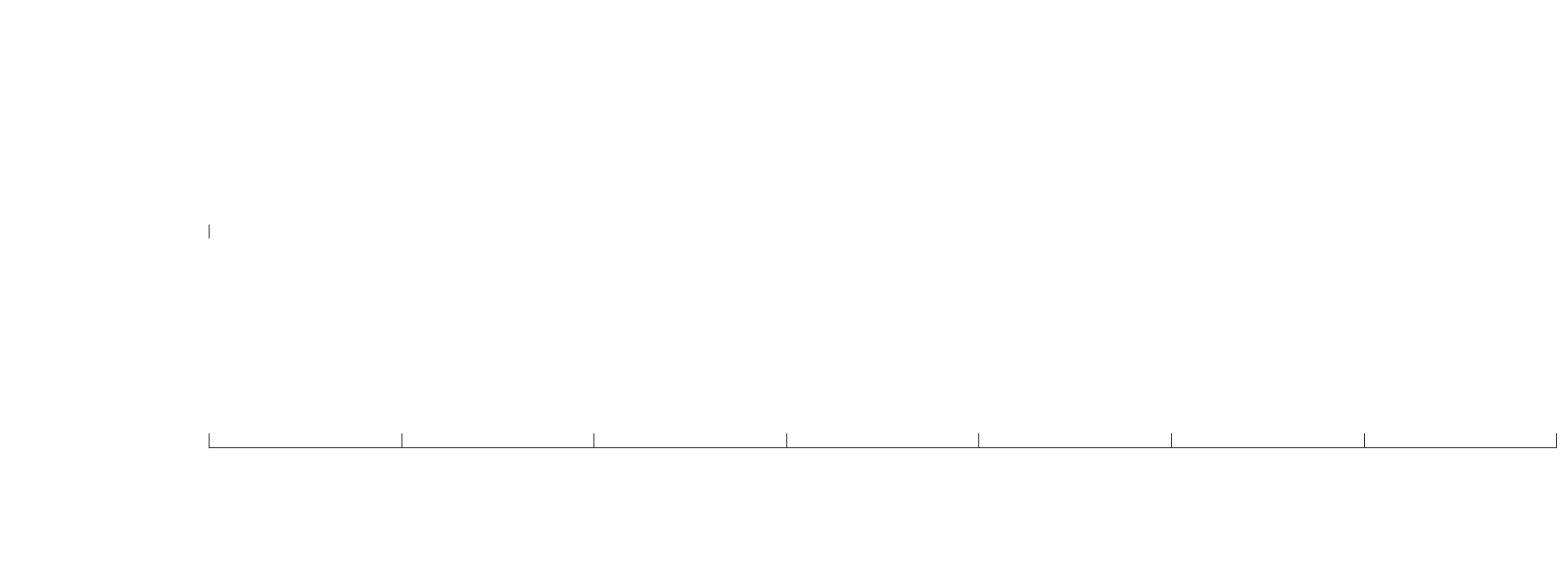}
\par\end{centering}
}
\par\end{centering}
\vspace{-5mm}

\begin{centering}
\subfloat[Square.]{\begin{centering}
\fontsize{25}{60}\resizebox{0.4\textwidth}{!}{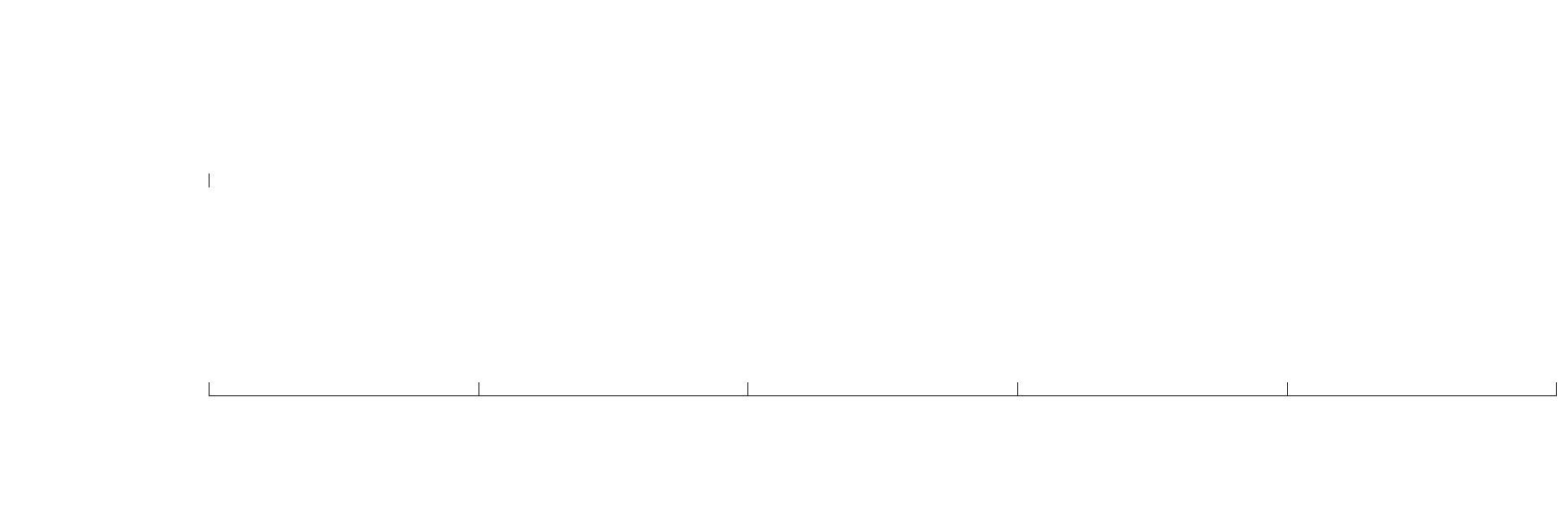}
\par\end{centering}
}
\par\end{centering}
\vspace{-5mm}

\begin{centering}
\subfloat[Triangle.]{\begin{centering}
\fontsize{27}{60}\resizebox{0.4\textwidth}{!}{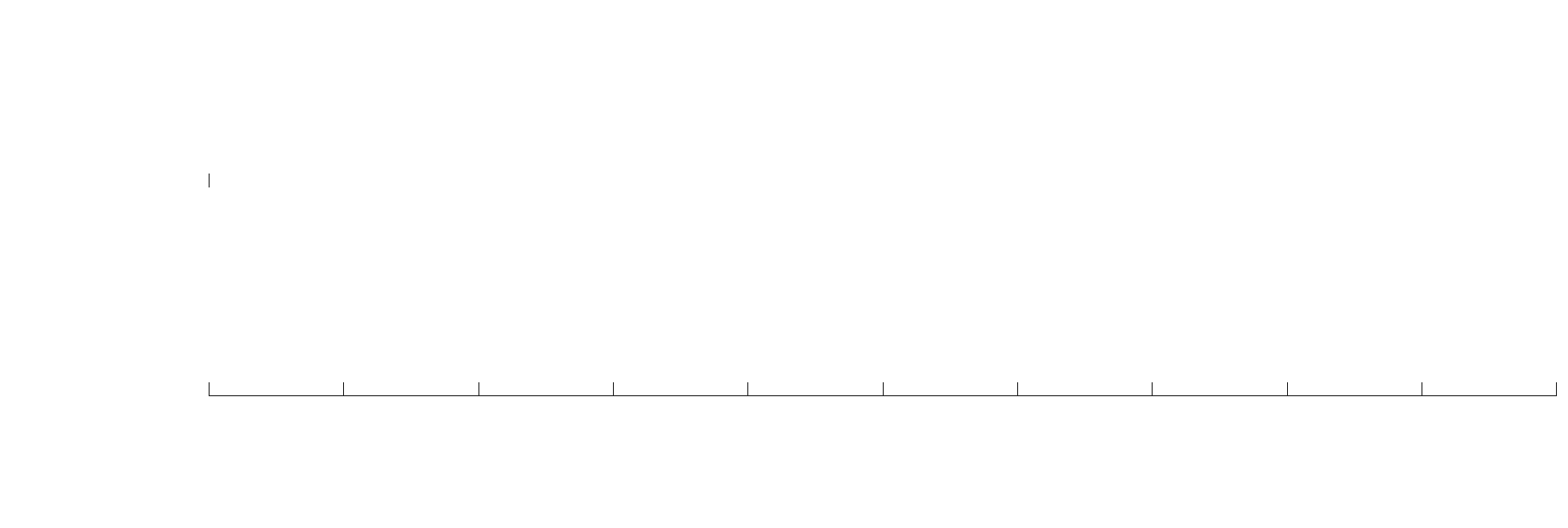}
\par\end{centering}
}
\par\end{centering}
\vspace{-5mm}

\begin{centering}
\subfloat[Diamond.]{\begin{centering}
\fontsize{25}{60}\resizebox{0.4\textwidth}{!}{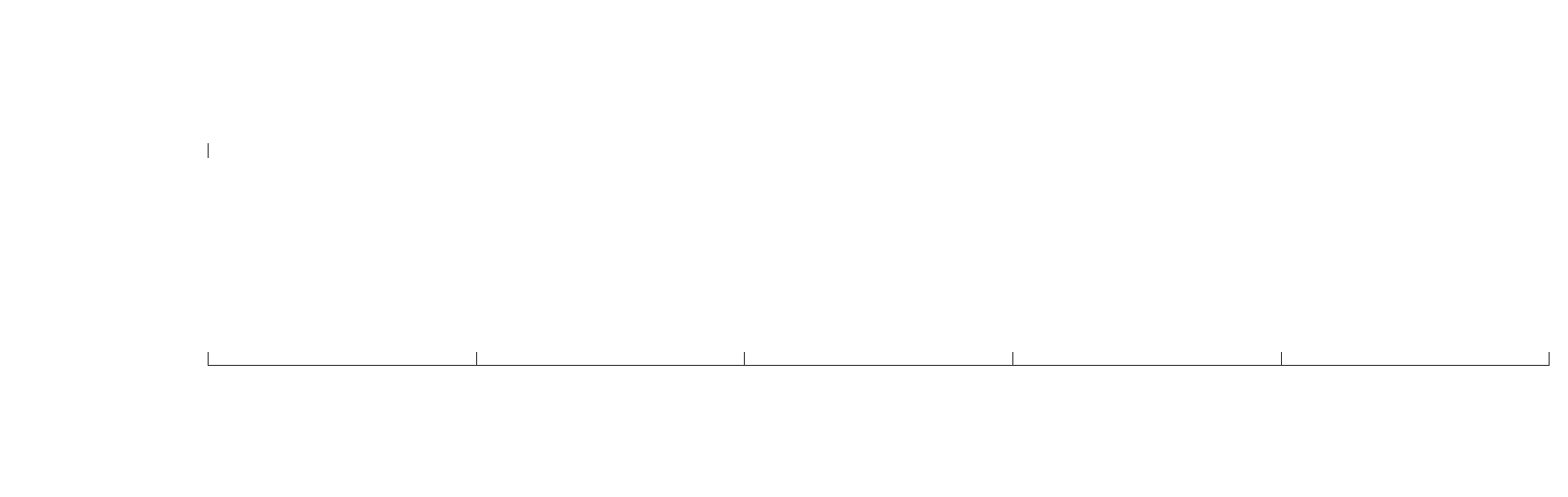}
\par\end{centering}
}
\par\end{centering}
\caption{Norm of the end-effector error while tracking different paths over
three circulations. The solid black line represents the norm of the
error for ACPO, while the blue dot-dashed line indicates the norm
of the error for ACLO. The vertical dashed lines mark the end of each
circulation over the cutting path.\label{fig:norm-error}\vspace{-5mm}
}
\end{figure}

Fig.~\ref{fig:end-effector-trajectories} presents the path traced
onto the photosensitive surface when using ACPO and ACLO for the paths
detailed in Section~\ref{subsec:cutting-path-definition}. When using
ACPO, the executed paths match most sections of the desired ones.
However, as shown in Fig.~\ref{fig:norm-error} (\emph{black} curve),
since the robot is completely actuated with respect to the task of
controlling the end-effector pose, activation of any of the 20 constraints
will conflict with the tracking task and prevent the error from converging
asymptotically to zero, resulting in deviations from the desired trajectory.
This can be seen in Fig.~\ref{fig:triangle-trajectory}, where the
red dashed line shifts away from the black line in some sections.
Additionally, the measurements for the adaptation law are obtained
from the visual system, which is noisy and induces fluctuations in
the estimated end-effector pose. This explains the high-frequency,
low amplitude (\textasciitilde 1~mm) oscillation in the time response.

On the other hand, relaxing the task to a line control objective induces
a functional redundancy that gives the controller more freedom to
minimize the objective function, and hence reduce the tracking error,
thereby compensating for activated constraints. As shown in Fig.~\ref{fig:norm-error}
(blue curve), having the flexibility of rotating around and translating
along the end-effector UV beam resulted in a smaller tracking error
for all tested paths.

Fig.~\ref{fig:norm-error} shows the error norm for tracking poses
with ACPO and lines with ACLO. Although from a control standpoint
those metrics are fundamental to assess whether the closed-loop system
works as intended, they do not provide an easily interpretable quantitative
metric for the tracking error because they involve norms in more abstract
spaces containing orientations (ACPO) and oriented lines (ACLO). Therefore,
even if the UV beam traced the nominal path accurately, the error
norms shown in Fig.~\ref{fig:norm-error} might have been affected
by larger orientation or line direction errors. Table~\ref{tab:tracing-accuracy}
summarizes the average accuracy of the \emph{traced path} ($\myvec p_{\mathrm{traced}}$)
with respect to the \emph{nominal} \emph{path} ($\myvec p_{\mathrm{nom}}$),
which is calculated as
\begin{align*}
\mathrm{accuracy} & =\frac{1}{T}\int^{T}_{0}\norm{\myvec p_{\mathrm{nom}}(t)-\myvec p_{\mathrm{traced}}(t)}dt.
\end{align*}
For ACPO, $\myvec p_{\mathrm{nom}}$ is obtained by finding the point
at which a line collinear with the $z$-axis of the desired end-effector
pose intersects the box surface plane, whereas for ACLO, we consider
the intersection between the reference line and the plane. Analogously,
to calculate $\myvec p_{\mathrm{traced}}$ we define a line collinear
with the UV beam, which is estimated using the end-effector pose measurements,
and find the point at which it intersects the box surface plane. Clearly,
ACLO renders a much more accurate system.

\begin{table}
\caption{Mean accuracy and standard deviation (sd), in mm, of the traced path.\label{tab:tracing-accuracy}}

\begingroup
\setlength{\tabcolsep}{3pt} 

\begin{tabular*}{1\columnwidth}{@{\extracolsep{\fill}}cccccc}
\hline 
 & Vertical line & Square & Triangle & Diamond & \textbf{Overall}\tabularnewline
\hline 
\hline 
ACPO & $3.5$ (sd $2.4$) & $3.2$ (sd $1.4$) & $5.9$ (sd $3.4$) & $3.3$ (sd $1.4$) & \textbf{3.9 (sd 2.5)}\tabularnewline
ACLO & $2.6$ (sd $1.3$) & $2.5$ (sd $1.5$) & $2.4$ (sd $1.2$) & $2.1$ (sd $1.1$) & \textbf{2.4 (sd 1.3)}\tabularnewline
\hline 
\end{tabular*}\endgroup
\end{table}

Nonetheless, it is important to note that the accuracy reported in
Table~\ref{tab:tracing-accuracy} measures how well the system tracks
the nominal path, but it does not indicate how close the nominal path
is to the desired cutting points on the physical box. Uncertainties
in the box location and other geometric parameters are the motivation
for using the adaptive control law (\ref{eq:adaptation_optimization_problem})
for both the robot and the box, as shown in Fig.~\ref{fig:system-diagram}.
As the estimated pose of the box converges to the measured one, we
can increase our confidence that the nominal path matches the desired
cutting points on the physical box, up to the measurement precision.
As the system currently lacks a ground-truth for the traced path on
the physical box, this is the best analysis for the closed-loop system
tracking accuracy we can perform.

\section{Conclusions\label{sec:Conclusions}}

This paper presented a low-cost mockup for laser-cutting tasks. It
consists of a cuboid container with UV-sensitive faces mounted on
a three-axis table and a $6$-DoF robotic manipulator equipped with
a UV torch that simulates the laser. The proposed setup uses a visual
system based on cameras and fiducial markers to provide an adaptive
constrained task-space controller with visual measurements to compensate
for inaccurate parameters, eliminating the need to calibrate the system.
Furthermore, the robot reactively avoids collisions with obstacles
while handling the UV torch since the controller explicitly accounts
for geometric constraints.

To enhance the laser-cutting path tracking, we control the UV beam
instead of the full end-effector pose. This reduces the required DoF
to perform the task from six to four and induces a functional redundancy
that gives the controller more freedom to minimize the objective function.
Experimental results have shown that, despite an initially uncalibrated
system, the system tracks different trajectories with an overall mean
accuracy of $3.9$ (sd $2.5$) mm when the end-effector pose is controlled
and $2.4$ (sd $1.3$) mm when the UV beam is controlled.

Future work will focus on replacing the visual perception based on
fiducial markers with a markerless system based on online 3D reconstruction.
We also plan to explore physics-based task sequencing strategies to
automate the cut path definition and demonstrate the system tracking
more complex trajectories. Lastly, we intend to deploy the laser-cutting
system, showcased in the mockup, in actual nuclear decommissioning.

\bibliographystyle{IEEEtran}
\bibliography{bibliography}

\end{document}